\documentclass[runningheads]{llncs}
\usepackage{graphicx}
\usepackage{color}

\usepackage{url}
\usepackage[colorlinks=true,linkcolor=blue, citecolor=blue, urlcolor=blue]{hyperref}
\usepackage{booktabs}
\usepackage{ulem}
\usepackage{amsmath}
\usepackage{subcaption}
\usepackage{xcolor,colortbl}
\definecolor{lavender}{gray}{0.9}
\usepackage{soul}
\usepackage{tabularx}
\usepackage{bbding}

\begin{document}
%
\title{Toro-SAM: Towards robust 3D interactive medical image segmentation with visual prompts}
\title{PIM: Towards robust 3D interactive medical image segmentation with visual prompts}
\title{PRISM: Towards robust 3D interactive medical image segmentation with visual prompts}
\title{PRISM: A \textbf{P}romptable and \textbf{R}obust \textbf{I}nteractive \textbf{S}egmentation \textbf{M}odel with Visual Prompts}
\titlerunning{PRISM}
%

\author{Hao Li \and Han Liu \and Dewei Hu \and Jiacheng Wang \and Ipek Oguz}
\authorrunning{H. Li et al.}
\institute{Vanderbilt University}
\maketitle              

\begin{abstract}
In this paper, we present PRISM, a \textbf{P}romptable and \textbf{R}obust \textbf{I}nteractive \textbf{S}egmentation \textbf{M}odel, aiming for precise segmentation of 3D medical images. PRISM accepts various visual inputs, including points, boxes, and scribbles as sparse prompts, as well as masks as dense prompts. Specifically, PRISM is designed with four principles to achieve robustness: (1) \underline{Iterative learning.} The model produces segmentations by using visual prompts from previous iterations to achieve progressive improvement. (2) \underline{Confidence learning.} PRISM employs multiple segmentation heads per input image, each generating a continuous map and a confidence score to optimize predictions. (3) \underline{Corrective learning.} Following each segmentation iteration, PRISM employs a shallow corrective refinement network to reassign mislabeled voxels. (4) \underline{Hybrid design.} PRISM integrates hybrid encoders to better capture both the local and global information. 
Comprehensive validation of PRISM is conducted using four public datasets for tumor segmentation in the colon, pancreas, liver, and kidney, highlighting challenges caused by anatomical variations and ambiguous boundaries in accurate tumor identification. Compared to state-of-the-art methods, both with and without prompt engineering, PRISM significantly improves performance, achieving results that are close to human levels. The code is publicly available at \url{https://github.com/MedICL-VU/PRISM}.
\keywords{Interactive medical image segmentation \and Iterative correction \and Tumor segmentation \and Visual prompts \and Segment Anything Model (SAM).}
\end{abstract}

\section{Introduction}
In recent years, deep learning-based methods have achieved state-of-the-art performance in various segmentation tasks \cite{liu2023medical,isensee2021nnu}. However, achieving robust outcomes remains a challenge due to significant anatomical variations among individuals and ambiguous boundaries in medical images. Alternatively, interactive segmentation models,
such as the Segment Anything Model (SAM) \cite{kirillov2023segment},
offer a solution by involving humans in the loop and utilizing their expertise to achieve precise segmentation. This requires users to indicate
the target region by providing visual prompts, such as points \cite{liu2023simpleclick,kirillov2023segment,luo2021mideepseg,gong20233dsam,li2023promise,li2023assessing,wang2018deepigeos,gao2023desam}, boxes \cite{wei2023medsam,zhang2023segment,deng2023sam,wang2018interactive,kirillov2023segment,ma2024segment,chen2023ma,yao2023false},
scribbles \cite{wang2018deepigeos,chen2023scribbleseg,wong2023scribbleprompt,cho2021deepscribble}, and masks \cite{kirillov2023segment,sun2023cfr,wang2023sam,cheng2023sam,liu2023simpleclick}. 

Importantly, a robust interactive segmentation model should be \underline{effective}
in responding to visual prompts given by users with minimal interactions. Inspired by SAM \cite{kirillov2023segment}, many interactive segmentation methods have been proposed in medical imaging
. However, most of these SAM-based methods are limited to using a single type of prompt \cite{wei2023medsam,gong20233dsam,yao2023false,gao2023desam,li2023promise,deng2023sam,ma2024segment,chen2023ma,zhang2023segment}. This limits the available information, impacting the effectiveness of the model. 
Due to the heterogeneity in anatomy and appearance of medical images, the models should leverage the advantages of different prompts to achieve optimal performance across various medical applications.
Moreover, these interactive methods do not involve humans in the loop \cite{gong20233dsam,gao2023desam,zhang2023segment,li2023promise,deng2023sam,wei2023medsam,ma2024segment,chen2023ma,yao2023false}, i.e., the visual prompts are applied to the automatic segmentation model only once, which may not be sufficient to achieve robust outcomes with the initial prompts provided. In contrast, practical applications often necessitate iterative corrections based on new prompts from the user until the outcome meets their criteria \cite{sofiiuk2021reviving}. Unfortunately, few studies have explored such human-in-loop approach for medical interactive segmentation \cite{cheng2023sam,wang2023sam}, and their performance has been suboptimal. 

An additional concern is the efficiency of the model with respect to training and user interaction requirements. Training a model from scratch with extensive datasets \cite{cheng2023sam,wong2023scribbleprompt,ma2024segment,wang2023sam} improves overall representational capabilities, but this is time-consuming and typically requires substantial computational resources for optimal performance. Furthermore, models may lack specificity for particular applications due to existing domain gaps in medical imaging \cite{wang2023sam}. This requires an increased number of user prompts to reach adequate performance. Moreover, 2D models \cite{cheng2023sam,wong2023scribbleprompt,ma2024segment} are not considered ``efficient'' for 3D images as they require significant user effort to provide visual prompts for each slice.

In this work, we propose a robust method for interactive segmentation in medical imaging. We strive for human-level performance, as a human-in-loop interactive segmentation model with prompts should gradually refine its outcomes until they closely match inter-rater variability \cite{haarburger2020radiomics}. We present PRISM, a \textbf{P}romptable and \textbf{R}obust \textbf{I}nteractive \textbf{S}egmentation \textbf{M}odel for 3D medical images, which accepts both sparse (points, boxes, scribbles) and dense (masks) visual prompts. We leverage an iterative learning \cite{kirillov2023segment,sun2023cfr} and sampling strategy \cite{sofiiuk2021reviving} to train PRISM to achieve continual improvements across iterations. The sparse visual prompts are sampled based on the erroneous region of the prediction from previous iteration, which simulates the human correcting behavior. For each iteration, multiple segmentation masks are generated \cite{li2018interactive} along with regressed confidence scores. The output with the highest score is selected, increasing the robustness of the model. This approach is similar to model ensembling \cite{linmans2023predictive}. Moreover, inspired from \cite{oguz2017gradient,jang2019interactive}, the output is fed into a shallow corrective refinement network to correct the mislabeled voxels and refine the final segmentation, adhering to the goal of being effectiveness. A hybrid image encoder with parallel convolutional and transformer paths \cite{li2022cats} is used as the backbone to better capture the local and global information from a medical image. Our main contributions are summarized as:
\begin{itemize}
    \item Interaction: PRISM accepts various visual prompts, offering versatility that effectively addresses the diverse challenges of medical imaging segmentation. This approach ensures precise outcomes through user-friendly interaction.
    
    \item Human-in-loop: PRISM is trained with iterative and confidence learning, allowing for continuous improvements and robust performance. With each user input, the performance progressively approaches expert-level  precision.
    
    \item Network architecture: PRISM employs a corrective refinement network and a hybrid encoder to produce precise segmentation for challenging conditions.

\end{itemize}
We evaluate PRISM on four public tumor datasets, including tumors in colon \cite{antonelli2022medical}, pancreas \cite{antonelli2022medical}, liver \cite{bilic2023liver} and kidney  \cite{heller2021state}, where  anatomical differences among individuals and ambiguous boundaries are present. Comprehensive validation is performed against state-of-the-art automatic and interactive  methods, and PRISM significantly outperforms all of them with substantial improvements.

\begin{figure}[t]
\centering
\includegraphics[width=\linewidth]{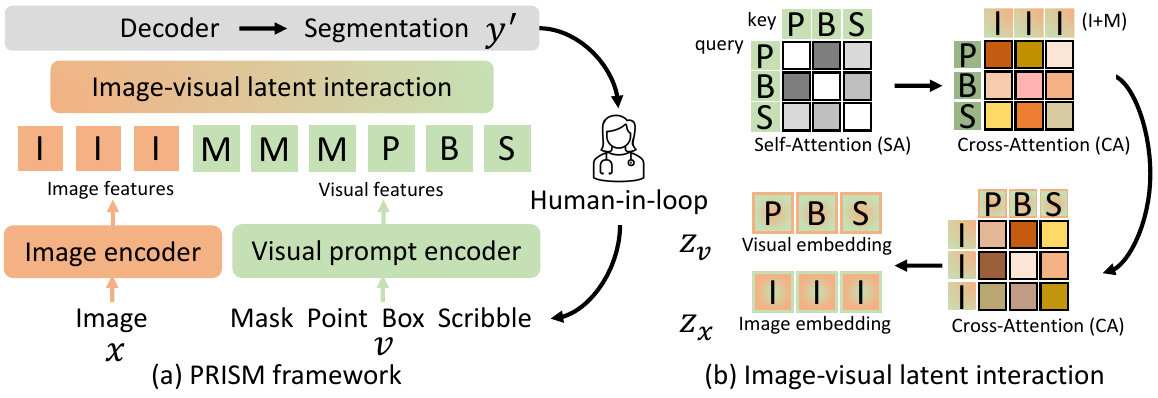}
\caption{\textbf{(a)} PRISM takes an image ($x$) and visual prompts ($v$) to produce a segmentation ($y')$. The user then provides prompts for next iteration. \textbf{(b)} Interaction between image and visual features in the latent space to produce image ($Z_x$) and visual ($Z_v$) embeddings with self- and cross-attention mechanisms.}
\label{PRISM}
\end{figure}

\section{Methods}
PRISM employs a generic encoder-decoder architecture as displayed in Fig.~\ref{PRISM}(a). For a given input image $x$, along with visual prompts $v$, the image and prompt encoders generate the image and visual features, respectively. These latent features then interact through self- and cross-attention mechanisms (Fig.~\ref{PRISM}(b)). The resulting embeddings, $Z_x$ and $Z_v$, are fed into the decoder to produce the final segmentation $y'$.  The expert then gives sparse prompt based on the erroneous regions in $y'$ for next iteration, in human-in-loop manner.

\begin{figure}[t]
\centering
\includegraphics[width=\linewidth]{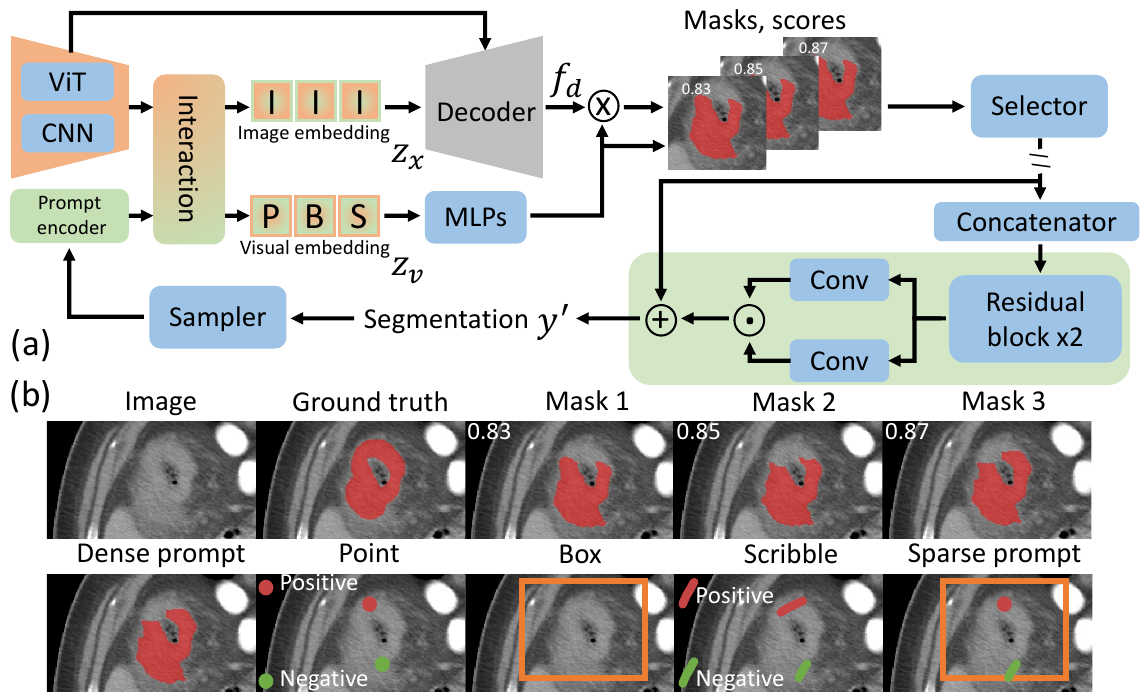}   
\caption{\textbf{(a)} Details of PRISM. Green highlights the corrective refinement network. \textbf{(b)} Top row shows the multi-mask prediction with labeled confidence scores. The selector would then pick Mask 3 as the dense prompt. Possible visual prompts given this dense prompt are shown in the bottom row. } 
\label{network_and_prompts}
\end{figure}

Fig.~\ref{network_and_prompts}(a) details PRISM, where a hybrid encoder combines parallel CNN and vision transformer paths to better extract image features. The last feature map of decoder $f_d$ is integrated with the visual embedding to generate multiple mask predictions and associated confidence scores (Fig.~\ref{network_and_prompts}(b)). A selector is employed to pick the candidate prediction with the highest confidence score, which is subsequently fed into the corrective refinement network to generate the $y'$. In our experiments, we use a sampler to mimic expert corrective actions by identifying point and scribble prompts from the false positive and negative areas within $y'$.

\noindent
\textbf{\textit{Iterative learning.}} PRISM is designed for clinical applications, which require  improvement through successive iterations.
We define the loss for a single iteration as $L_i$, and the total loss as $L_{total} = \sum_{i=1}^{N} L_i$, where $N$ denotes the total number of iterations. The input visual prompts come only from the current iteration, i.e., they are not cumulative. Since the dense prompt $y'_i$ is generated from the previous iteration and retains the gradient, PRISM can learn the relationship between iterations. For efficiency, the image features are generated only once at the initial iteration.
The following prompt types are allowed (Fig.~\ref{network_and_prompts}(b)):

\begin{itemize}
\item At each iteration $i$, point prompts are randomly sampled with uniform distribution from the false negative (FN) and positive (FP) regions of $y'_{i-1}$.
\item The scribbles are generated similar to Scribbleprompt \cite{wong2023scribbleprompt} by first extracting the skeleton of the FN and FP regions of $y'_{i-1}$ from the previous iteration. Next, a random mask is created by pixels randomly sampled from a normal distribution, applying Gaussian blur, and thresholding the image based on its mean value. This random mask is applied to divide the skeleton of $y'_{i-1}$ into separate, smaller parts. Finally, a random deformation field and random Gaussian filtering are used to change the scribble curvature and thickness. 
\item The 3D bounding box (BB) is determined based on the ground truth $y$ at the initial iteration ($i=1$) and remains unchanged for the subsequent iterations. 
\item The logits map at iteration $i-1$ is used to provide additional information instead of a binary mask as the dense prompt for iteration $i$, for $i>1$.
\end{itemize}



\noindent
\textbf{\textit{Confidence learning.}} Unlike traditional methods that output a single mask, PRISM increases robustness by generating multiple outputs, each with a confidence score. The last feature map of the decoder, denoted as $f_d=Decoder(Z_x)$, interacts with the visual embedding to produce continuous maps $m_j$ and confidence scores $s_j$ via $m_{j} = f_d \times MLP^m_{j}(Z_v)$ and $s_{j} = MLP^s_{j}(Z_v)$, where $MLP$ indicates the multi-layer perceptrons, and $j$ is an index over the multiple outputs. Thus, $m_{j}$  interacts with visual prompts at both latent- and image-level. The process of confidence supervision for each iteration is defined by: $L_{con} = \sum_{j=1}^{M}( \lambda_s L_s(m_{j}, y) + \lambda_b L_b(m_{j}, y) + \lambda_r L_r(s_{j}, 1-L_{Dice}(m_{j}, y)))$, where $L_s$, $L_b$, $L_r$ and $L_{Dice}$ indicate structural, boundary, regression and Dice loss, respectively. More precisely, the structural information is captured through a combination of Dice loss and cross-entropy loss: $L_s = L_{Dice} + L_{CE}$. For a given logits map or binary mask, the smooth boundary is derived as: $B(m) = |m - P_{ave}(m)|$, where $P_{ave}$ denotes the average pooling layer. The boundary loss ($L_b$) is evaluated by $L_{MSE}(B(m), B(y))$, with $L_{MSE}$ representing the mean square error loss. We also use $L_{MSE}$ for the regression loss. 


\noindent
\textbf{\textit{Corrective learning.}} As shown in Fig.~\ref{network_and_prompts}(a), a selector is employed to choose the candidate mask $\hat{m}$ corresponding to the highest confidence score $s_j$, which is then fed into the corrective refinement network. We design this network to be both shallow and efficient, consisting of two cascaded residual blocks and two convolution operations in two different branches. It takes the input $x_c$ at original image resolution. Importantly, $x_c$ is the concatenation of 4 images: (1) the input image $x$, (2) the binary mask $\hat{b}$ derived from $\hat{m}$, 
(3) the binary positive prompt maps accumulated from iteration 0 to i, and (4) the similar binary cumulative negative prompt map. 
For efficiency, the $x_c$ is downsampled and then upsampled back for $y'_i$.
The final segmentation $y'_i$ is supervised by: $L_{cor} = \lambda_s L_s(y'_i, y) + \lambda_b L_b(y'_i, y)$. Notably, the corrective network does not backpropagate to the network weights for candidate mask generation and focuses exclusively on refining any given logit map. The $y'_i$ is fed to sampler for next iteration $i+1$.

\noindent
\textbf{\textit{Implementation details.}}
For training, the total objective function is: $L_{total} = \sum_{i=1}^{N} (L_{{con}_i} + L_{{cor}_i})$, 
where the ratio of $\lambda_s : \lambda_b : \lambda_r$ as $1: 10: 1$. The batch size was set to 2, and the initial learning rate was $4e^{-5}$, which was reduced by $2e^{-6}$ after every epoch. We employed the AdamW optimizer across a maximum of 200 epochs. Our study was conducted on an NVIDIA A6000. The input image was cropped to a 3D $128\times128\times128$ patch centered on a foreground pixel. The  number of iterations ($N$) was fixed at 11, with 1 to 50 point prompts randomly chosen per iteration during training. The Dice and normalized surface Dice (NSD) are used as evaluation metrics. We compared to both fully automated \cite{isensee2021nnu,tang2022selfsupervised,lee2023d} and interactive \cite{gong20233dsam,li2023promise,wang2023sam,kirillov2023segment} methods. We used $y'_{i=11}$  as final segmentation for the iterative methods \cite{wang2023sam} and PRISM. We retrained using the published code of each model, and the pretrained weights were also employed if publicly available. We present two versions of our method. PRISM-plain only uses point prompts, while PRISM-ultra can handle other sparse visual prompts such as boxes and scribbles. All results are generated with seeds.








\section{Experiments}
\noindent
\textbf{\textit{Datasets.}}
We use four public 3D CT datasets, including colon (N=126) \cite{antonelli2022medical}, pancreas (N=281) \cite{antonelli2022medical}, liver (N=118) \cite{bilic2023liver} and kidney  (N=300) \cite{heller2021state} tumors. Challenges include large anatomical differences among individuals, varying shapes of target structures, and ambiguous boundaries. Additionally, multiple tumors \cite{heller2021state,bilic2023liver} may occur in a single subject.  We adopted the same data split as a prior study \cite{gong20233dsam} for each task with a training/validation/testing split of 0.7/0.1/0.2, and we only use tumor labels to focus on binary segmentation. We resampled to a $1mm$ isotropic resolution, performed intensity clipping based on the $0.5$ and $99.5$ percentiles of the foreground, and applied Z-score normalization based on foreground voxels. Random zoom and intensity shift are used as data augmentations.

\begin{table*}[t]
\caption{Quantitative results. Bold  indicates best performance.  PRISM-plain is a simplified model for a fair comparison to methods not allowing BB prompts \cite{gong20233dsam,li2023promise,wang2023sam}, and methods\cite{kirillov2023segment,gong20233dsam,li2023promise,wang2023sam} only using 1 point prompt per input. For liver tumor, 10 points are used for all methods to accommodate multiple tumors.} 

\label{main table}
\small
\begin{center}
    \begin{tabular}{ l  | c |c |c |c}
    \toprule
    \multicolumn{1}{c}{ } & \multicolumn{4}{c}{Public datastes (Dice $\%$ / NSD $\%$)} \\ 
    \hline
    \multicolumn{1}{l}{Methods} &   \multicolumn{1}{c}{Colon tumor} & \multicolumn{1}{c}{Pancreas tumor} & \multicolumn{1}{c}{Liver tumor} & \multicolumn{1}{c}{Kidney tumor} \\
    \hline
    nnU-Net \cite{isensee2021nnu} &    43.91 / 52.52 & 41.65 / 62.54 & 60.10 / 75.41 & 73.07 / 77.47 \\

    3D UX-Net \cite{lee2023d}  &  28.50 / 32.73 & 34.83 / 52.56 & 45.54 / 60.67 & 57.59 / 58.55 \\


    Swin-UNETR \cite{tang2022selfsupervised} &   35.21 / 42.94 & 40.57 / 60.05 & 50.26 / 64.32 & 65.54 / 72.04 \\
    \hline
    
    SAM \cite{kirillov2023segment} &   28.83 / 33.63 & 24.01 / 26.74 & 8.56 / 5.97 & 36.30 / 29.86\\ 


    3DSAM-adapter \cite{gong20233dsam} &   57.32 / 73.65 & 54.41 / 77.88 & 56.61 / 69.52 & 73.78 / 83.86 \\

    ProMISe \cite{li2023promise} &   66.81 / 81.24 & 57.46 / 79.76 &  58.78 / 71.52 &  75.70 / 80.08 \\

    SAM-Med3D \cite{wang2023sam} &  54.34 / 78.58  &  65.61 / 92.40 & 23.64 / 26.97  &  76.50 / 88.41 \\
    SAM-Med3D-organ &   70.75 / 91.03  & 76.40 / 97.75  &  66.52 / 77.97  & 88.20 / 97.80 \\
    SAM-Med3D-turbo &   73.77 / 94.95  & 74.87 / 96.43  & 69.36 / 81.70 &  89.26 / 98.40 \\

    \hline 
    PRISM-plain &  67.18 / 85.28   &  65.73 / 89.51 &  79.70 / 91.60  &  85.29 / 93.55 \\

    PRISM-ultra &  \textbf{93.79} / \textbf{99.96}  &  \textbf{94.48} / \textbf{99.99} &  \textbf{94.18} / \textbf{99.99} &  \textbf{96.58} / \textbf{99.80} \\
    
    \hline     

\multicolumn{5}{l}{
\begin{tabular}{@{}l@{}@{}} 
    ``PRISM-plain'' only uses 1 point and no BB. 
    ``-ultra'' adds the BB, and scribbles. 
\end{tabular}
    } 
     
\end{tabular} 

\end{center}
\end{table*}

\noindent
\textbf{\textit{Comparison with state-of-the-art.}} Table \ref{main table} compares quantitative segmentation results (See qualitative results in Fig. \ref{quali_supp}). As expected, fully automated methods \cite{isensee2021nnu,tang2022selfsupervised,lee2023d} are unable to deliver satisfactory outcomes for these challenging datasets. Even with visual prompts, SAM \cite{kirillov2023segment} struggles with the substantial domain shift between natural and medical images. Furthermore, the interactive segmentation models \cite{gong20233dsam,li2023promise,wang2023sam} trained from scratch fail to produce adequate results. Notably, ``-organ'' and ``-turbo'' denote the fine-tuning of the pretrained SAM-Med3D model \cite{wang2023sam}. ``-organ'' focuses on organ-specific datasets, while ``-turbo'' includes these organ datasets (including, in some cases, the test data in their pretraining) alongside a broader range of other medical imaging data. In contrast, excluding the pretrained models, the proposed PRISM-plain demonstrates superior overall Dice performance with only a single  point as sparse prompt. Moreover, as more visual prompts (points, BB, and scribbles) are provided, PRISM-ultra delivers outcomes close to human level performance, demonstrating its strength as an effective human-in-loop algorithm.

\begin{figure}[t]
\centering
\includegraphics[width=\linewidth]{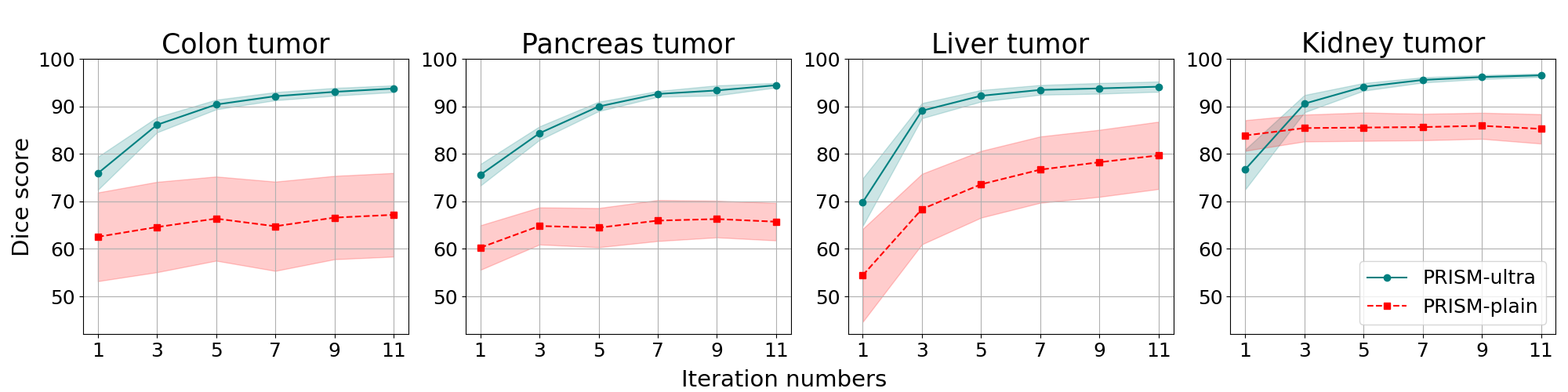}   
\caption{Dice score of proposed PRISM on four tumor datasets, where the mean values (lines) and their 95\% confidence intervals (shades) are presented.}
\label{iter_plot}
\end{figure}

\begin{figure}[t]
\centering
\includegraphics[width=\linewidth]{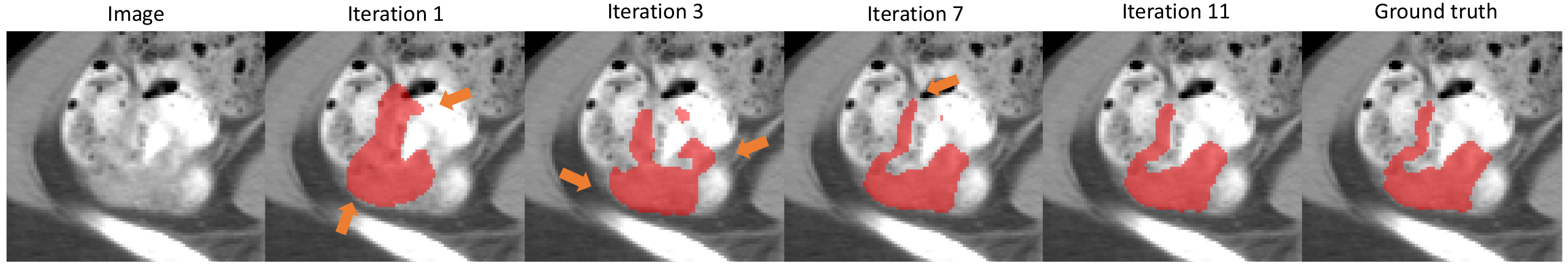}   
\caption{Qualitative results of PRISM-ultra for colon tumors characterized by irregular shapes and ambiguous boundaries. The orange arrows indicate the major defects which are corrected in the subsequent iteration. The initial output has noticeable errors that rapidly get corrected in the first few iterations. More qualitative results can be viewed in Fig.~\ref{iter_supp}.}
\label{qualitative_iter_results}
\end{figure}

\noindent
\textbf{\textit{Analysis of iterative learning performance.}} Fig.~\ref{iter_plot} shows the performance of PRISM across the iterative learning process. Although PRISM-plain delivers superior outcomes (Table \ref{main table}),  it does not have monotonically increasing performance across iterations and can suffer from a decrease in performance at later iterations, except for liver tumor. In practice this would be frustrating to the human user, as more input prompts paradoxically lowers the segmentation accuracy. However, PRISM-ultra presents not just substantial improvements in overall performance, but also a monotonically increasing accuracy and narrower 95\% confidence intervals. This highlights the PRISM-ultra as a robust model, and its qualitative results in Fig.~\ref{qualitative_iter_results} and Fig.~\ref{iter_supp} illustrate the iterative correction. 
Initially, the results may not meet expectations due to ambiguous boundaries and anatomical variations. Yet, as iterations advance, the outcomes progressively align more closely with the human label. Major corrections occur at the early stages which indicate the efficiency and effectiveness of PRISM-ultra.

\begin{table*}[t]
\caption{Ablation study on the colon tumor. ConL: confidence learning. PRISM-plain-b: PRISM-plain plus BB. Model configurations are detailed in Table \ref{architecture_setting_table}.}
\label{ablation table}
\sethlcolor{gray!20} 
\begin{center}
    \begin{tabular}{ l  | c |c |c |c | c | c}
    \toprule
     & ViT encoder & CNN encoder & Hybrid & PRISM-plain & PRISM-plain-b & no ConL \\

    \hline
    Dice & 58.38  & 63.58  & 66.40 & 67.18  &  76.40 &  73.65 \\

    \hline     
     
  \end{tabular} 
\end{center}

\caption{Prompt analysis for colon tumors. ``-1'', ``-50'': number of points used during training. PRISM-basic: PRISM-plain-b with random number of points. ``-erode'' and ``-dilate'':  modified BB by 5 voxel radius. CorL:  corrective learning. PRISM-ultra: PRISM-basic plus test-time scribbles. PRISM-ultra+: PRISM-ultra plus train-time scribbles. Prompt configurations are detailed in Table \ref{prompt_setting_table}.}

\label{point table}

\begin{center}
    \begin{tabular}{ c | c |c |c | c |c | c | c | c}
    \toprule
    Test Points & -plain-b-1 & -plain-b-50 & -basic & no CorL & \;-erode\; &  \;-dilate\; & \;-ultra\; & \;-ultra+\\

    \hline
    1   &   76.40  &  61.33  &  76.66  & 77.65  &  72.99  & 75.47  & 93.79 & 96.47 \\

    10  &  77.71   &  82.42  &  83.85  & 80.12 &  82.85  & 83.72  & 93.95 & 96.54 \\
    50  &  72.52   &  89.91  &  89.36  & 81.60  &  89.99  & 89.42  & 94.07 &  96.73 \\
    100 &  69.60   &  91.30  &  90.26  & 81.20  &  91.06  & 90.70  & 94.28  &  96.93 \\

    \hline     
     
  \end{tabular} 
\end{center}


\end{table*}


\noindent
\textbf{\textit{Analysis of framework.}} 
An ablation study for various frameworks on the colon dataset are presented in Table \ref{ablation table}, and the details of model configurations in Table~\ref{architecture_setting_table}. The CNN encoder outperforms the use of a Vision Transformer (ViT) alone. Using a hybrid encoder improves results, while proposed confidence and corrective learning further increase PRISM-plain performance. PRISM-plain is advanced to PRISM-plain-b through the incorporation of a 3D BB, a favored prompt in medical segmentation due to its strong image prior, leading to improved outcomes. For efficiency, drawing a box does not require significantly more effort than point prompt. For effectiveness, points can address the limited flexibility of boxes in making corrections in subsequent iterations. The  proposed confidence learning approach also contributes to the performance of the model.

\noindent
\textbf{\textit{Analysis of number of prompts.}} We analyze the performance with respect to the number of point prompts in Table \ref{point table}. The compared model configurations are detailed in Table \ref{prompt_setting_table}. We first evaluate the number of prompts used for training. The model plain-b-1 uses 1 prompt for training, plain-b-50 uses 50, and basic uses a random number between 1 and 50. Too few or too many training points lowers performance, while the random selection is the most robust. We also note that  corrective learning has a more pronounced contribution when more prompts are used. 
We next test the sensitivity to BB precision by dilating and eroding the true BB. Either eroding or dilating BB decreases performance compared to PRISM-basic when few points are used, but using more points resolves this issue.
Adding test-time scribbles (ultra) and train-time scribbles (ultra+) both improves the performance of PRISM-basic. However, this improvement increases training time, making scribbles less practical for large datasets. 




\section{Conclusion}
In this study, we propose PRISM, a \textbf{P}romptable \textbf{R}obust \textbf{I}nteractive \textbf{S}egmentation \textbf{M}odel for 3D medical images. It supports various visual prompts and applies diverse learning strategies to achieve performance comparable to that of human raters. This is demonstrated across four challenging tumor segmentation tasks, where existing state-of-the-art automatic and interactive segmentation approaches do not meet the expected standards. Future work will include   adaptation of pretrained  models, and learning from datasets with sparse annotations.

\noindent
\textbf{\textit{Acknowledgments.}}
This work was supported, in part, by NIH U01-NS106845, and NSF grant 2220401.

\appendix
\renewcommand\thefigure{S\thesection.\arabic{figure}}\renewcommand\thetable{S\thesection.\arabic{table}}
\renewcommand\thesubsection{S\thesection.\arabic{subsection}}

%
%
%

\bibliographystyle{splncs04}
\bibliography{mybibliography}
%

\clearpage
\setcounter{secnumdepth}{0}
\begin{center}
{ \textbf{Supplementary Materials}} 
\end{center}
\begin{figure}[!ht]
\centering
\setcounter{figure}{0}
\includegraphics[width=0.9\linewidth]{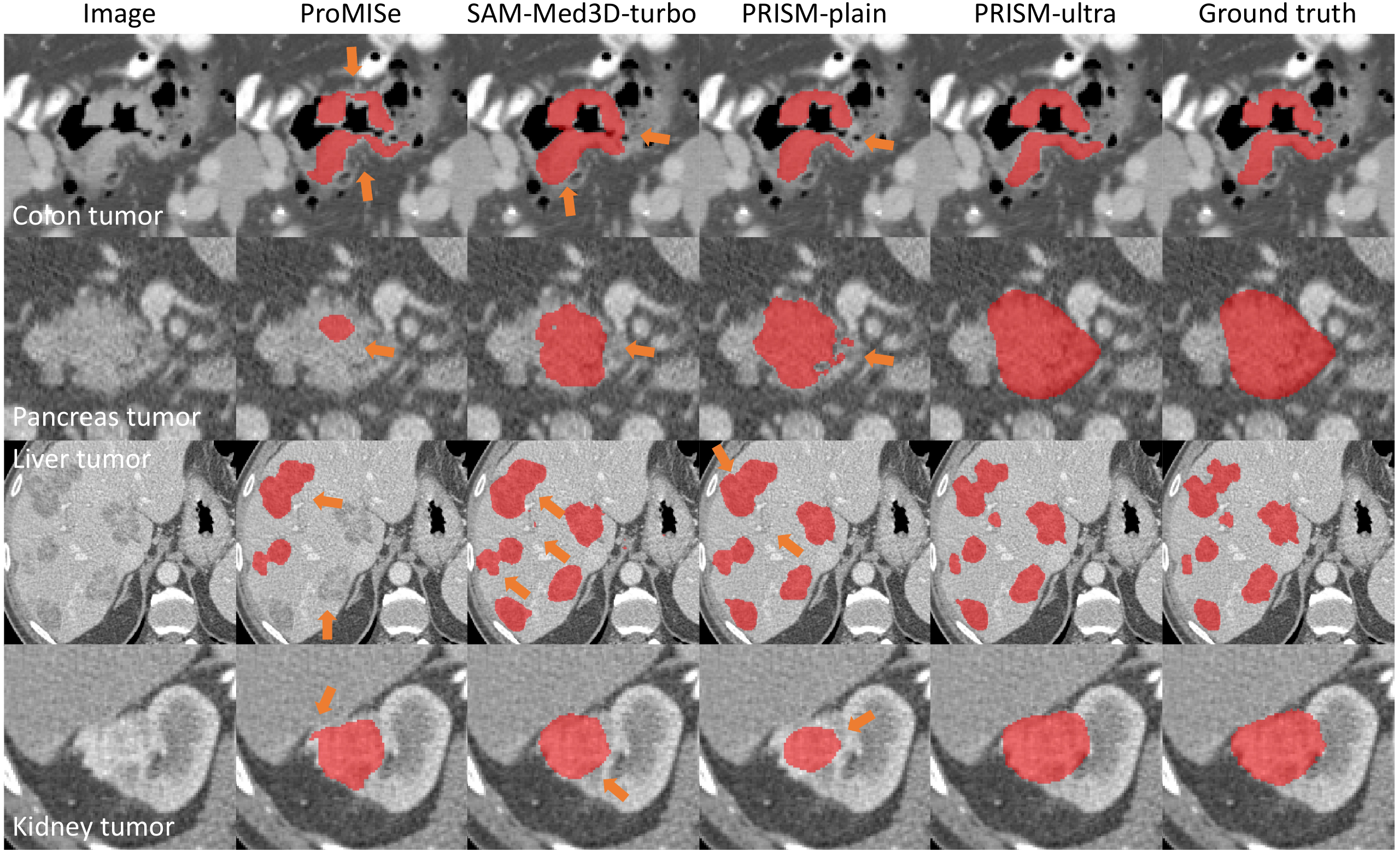}   
\caption{Qualitative results of four different tumor segmentation tasks. The orange arrows indicate the major defects.} 
\label{quali_supp}
\end{figure}

\begin{figure}[!ht]
\centering
\includegraphics[width=0.9\linewidth]{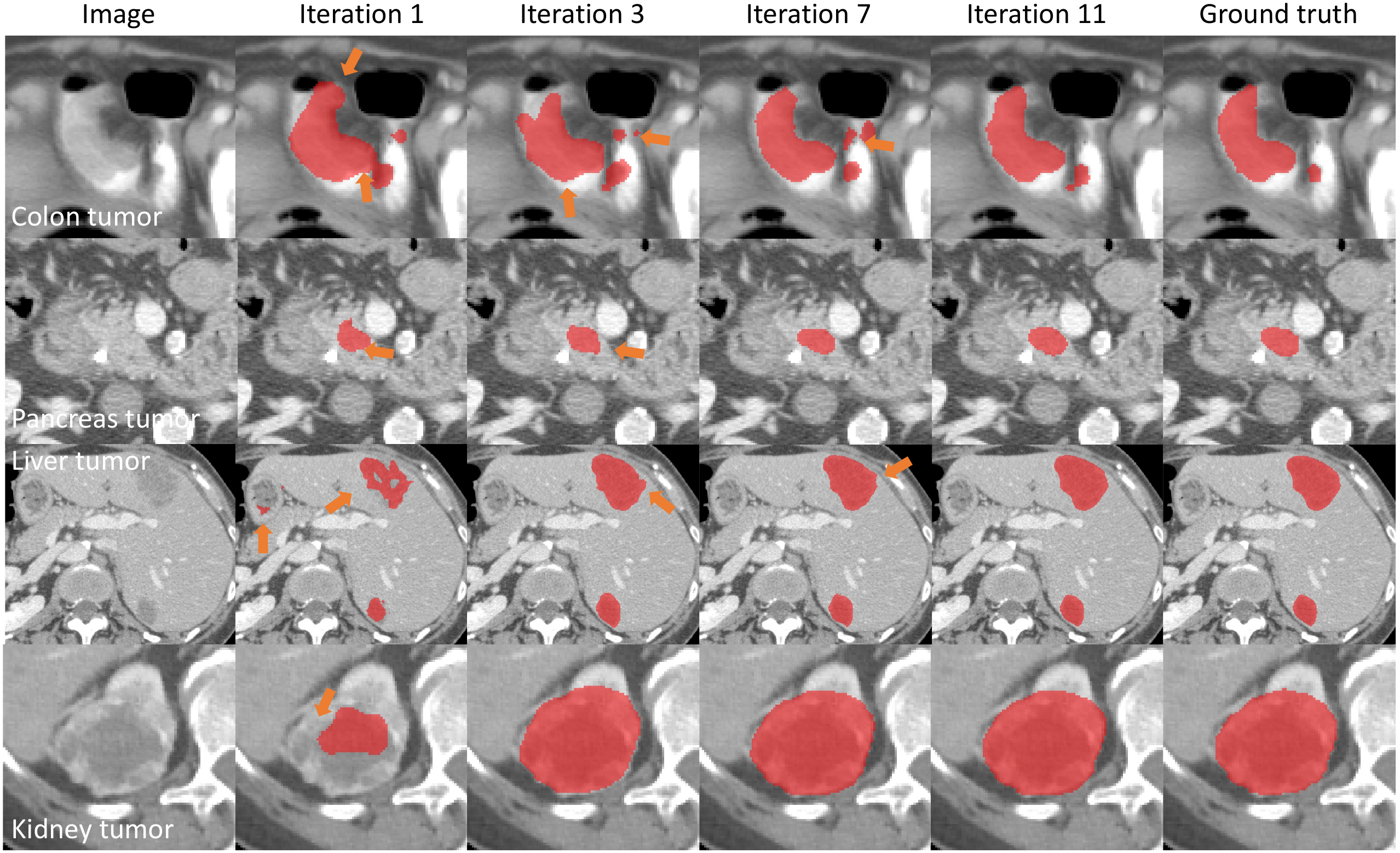}   
\caption{Qualitative results of PRISM-ultra across iterations. The orange arrows indicate the major defects which are corrected in the subsequent iteration.} 
\label{iter_supp}
\end{figure}

\setcounter{secnumdepth}{2}
\setcounter{figure}{0} 
\setcounter{table}{0} 
\begin{table*}[h]
\caption{Detailed settings for architecture and learning strategy in the proposed ablation studies.}
\label{architecture_setting_table}
\begin{center}
    \begin{tabular}{ l  | c |c |c |c | c}
    \toprule
     \multicolumn{1}{c}{ } & \multicolumn{3}{c}{Encoder type} & \multicolumn{2}{c}{Learning strategy}\\ 
    \hline
    Variants & ViT & CNN & ViT + CNN & Confidence learning & Corrective learning \\    
    \hline
    ViT encoder   & \Checkmark   &                 &    &   &\\
    CNN encoder   &              &  \Checkmark     &    &   &\\
    Hybrid        &              &                 &  \Checkmark  &   &\\
    PRISM-plain   &              &                 &  \Checkmark  &   \Checkmark &  \Checkmark \\
    PRISM-plain-b &              &                 &  \Checkmark  &   \Checkmark &  \Checkmark \\
    no ConL       &              &                 &  \Checkmark  &              & \Checkmark\\

    PRISM-basic   &    &         &  \Checkmark  &  \Checkmark & \Checkmark\\
    no CorL       &    &         &  \Checkmark  &  \Checkmark &\\
  
    PRISM-ultra        &    &    &  \Checkmark  &  \Checkmark & \Checkmark\\

    \hline       
  \end{tabular} 
\end{center}


\caption{Detailed prompt setting for the proposed ablation studies. Tr. and T. represent training and test, respectively. The notation [1, 50] indicates the range from which the number of point prompts is randomly sampled during training. The term ``varies'' refers to the number of points used in tests, which include 1, 10, 50, and 100. ``-erode'' and ``-dilate'' denote box sizes that are 5 voxels smaller or larger in each dimension, respectively.}
\label{prompt_setting_table}

\begin{center}
    \begin{tabular}{ l  | c |c |c |c | c}
    \toprule
     \multicolumn{1}{c}{ } & \multicolumn{5}{c}{Detailed prompt setting} \\ 
    \hline
    Variants & point & point num. (Tr. / T.) & box &  box type in T. &  scribble usage \\
    \hline
    ViT encoder   & \Checkmark   &  1 / 1             &    &   &\\
    CNN encoder   & \Checkmark   &  1 / 1             &    &   &\\
    PRISM-plain   & \Checkmark   &  1 / 1             &    &   &\\
    PRISM-plain-b & \Checkmark   &  1 / 1             &  \Checkmark  &  tight &\\
    no ConL       & \Checkmark   &  1 / 1             &  \Checkmark  &  tight &\\
    -plain-b-1    & \Checkmark   &  1 / 1             &  \Checkmark  &  tight &\\
    -plain-b-50   & \Checkmark   &  50 / varies       &  \Checkmark  &  tight &\\
    PRISM-basic   & \Checkmark   & [1, 50] / 1        &  \Checkmark  &  tight &\\
    no CorL       & \Checkmark   & [1, 50] / 1        &  \Checkmark  &  tight &\\
    -erode        & \Checkmark   & [1, 50] / varies   &  \Checkmark  &  undersized &\\
    -dilate         & \Checkmark   & [1, 50] / varies   &  \Checkmark  &  oversized &\\
    PRISM-ultra        & \Checkmark   & [1, 50] / varies   &  \Checkmark  &  tight & T.\\
    PRISM-ultra+       & \Checkmark   & [1, 50] / varies   &  \Checkmark  &  tight & Tr. and T.\\

    \hline       
  \end{tabular} 
\end{center}

\end{table*}

\end{document}